\documentclass{article}
\usepackage{ijcai20}

\usepackage{times}
\usepackage{soul}
\usepackage{url}
\usepackage[hidelinks]{hyperref}
\usepackage[utf8]{inputenc}
\usepackage[small]{caption}
\usepackage{graphicx}
\usepackage{amsmath}
\usepackage{amsthm}
\usepackage{booktabs}
\usepackage{algorithm}
\usepackage{algorithmic}
\usepackage{amsfonts}
\usepackage{float}

\usepackage{xcolor}

\newtheorem{theorem}{\textbf{Theorem}}

\newcommand{\D}{\mathcal{D}}
\newcommand{\Proba}{\mathbb{P}}
\newcommand{\R}{\mathbb{R}}

\newcommand{\E}{\mathbb{E}}

\newcommand{\x}{\mathbf{x}}

\newcommand{\y}{\mathbf{y}}

\newcommand{\balpha}{\boldsymbol{\alpha}}

\newcommand{\calX}{\mathcal{X}}

\title{Discriminative Active Learning for Domain Adaptation}

\author{
Fan Zhou$^1$\footnote{Corresponding author. Preprint, work in progress.}
\and
Changjian Shui$^1$\and
Bincheng Huang$^{2}$\and
Boyu Wang$^{3}$\And
Brahim Chaib-draa$^1$
\affiliations
$^1$Université Laval \qquad $^2$China Electronics Technology Group \qquad 
$^3$University of Western Ontario 
}

\begin{document}

\maketitle

\begin{abstract}
Domain Adaptation aiming to learn a transferable feature between different but related domains has been well investigated and has shown excellent empirical performances. Previous works mainly focused on matching the marginal feature distributions using the adversarial training methods while assuming the conditional relations between the source and target domain remained unchanged, $i.e.$, ignoring the conditional shift problem. However, recent works have shown that such a conditional shift problem exists and can hinder the adaptation process. To address this issue, we have to leverage labeled data from the target domain, but collecting labeled data can be quite expensive and time-consuming. To this end, we introduce a discriminative active learning approach for domain adaptation to reduce the efforts of data annotation. Specifically, we propose three-stage active adversarial training of neural networks: invariant feature space learning (first stage), uncertainty and diversity criteria and their trade-off for query strategy (second stage) and re-training with queried target labels (third stage).  Empirical comparisons with existing domain adaptation methods using four benchmark datasets demonstrate the effectiveness of the proposed approach.

\end{abstract}

\section{Introduction}

 
In general machine learning tasks, we usually assume the datasets, where the hypothesis was trained and tested, are from the same distribution.
However, this assumption, in general, is not realistic in many practical scenarios. For example, appearance shifts caused by illumination, seasonal, or weather changes are significant challenges for computer vision-based systems. A vision system trained on one dataset but deployed on another may suffer from rapid performance drop. 
More severely, to train a high-performance vision system requires a large amount of labeled data, and getting such labels may be expensive. 
One approach to deal with this issue is \emph{Domain Adaptation} (DA), which aims to improve the learning performance of a target domain by leveraging the unlabeled data in the target domain as well as the labeled data from a different but related domain (source domain).
Previous works have theoretically analyzed the learning guarantees of DA \cite{ben2010theory,redko2017theoretical} and have reported some empirical applications in natural language processing \cite{glorot2011domain} and computer vision \cite{wang2018visual}. 

Most recent DA advancements are mostly based on the basic \emph{Covariate Shift} assumption that the marginal distributions of source and target domain change $(\Proba_S(\x)\neq \Proba_T(\x))$ while the conditional distribution (predictive relation) is preserved $(\Proba_S(y|\mathbf{x}) = \Proba_T(y|\mathbf{x}))$ during the adaptation process. However, some recent works have revealed that this assumption may not hold, and in this case, one may still need some labeled data from the target domain in order to successfully transfer information from one domain to another. Specifically, \cite{zhao2019learning} discussed the conditional shift problem showing that such a problem exists and can hinder the adaptation process. They proved that the risk on target domain is controlled by the source risk, the marginal distribution divergence, and disagreement between the \emph{two labeling distributions}:
  \begin{equation}\label{Eq: Lemma from Han Zhao}
        \begin{split}
            \epsilon_{\mathcal{T}}(h)\leq
            &\epsilon_\mathcal{S}(h)+d_{\hat{\mathcal{H}}}(\mathcal{D}_{\mathcal{S}},\mathcal{D}_{\mathcal{T}}) \\
            &+\underbrace{\min\{\mathbb{E}_{\mathcal{D}_\mathcal{S}}\big[\big|f_\mathcal{S}-f_\mathcal{T}\big|\big],\mathbb{E}_{\mathcal{D}_\mathcal{T}}\big[\big|f_\mathcal{S}-f_\mathcal{T}\big|\big]\}}_{\text{Impossible to measure in unsupervised DA}}
        \end{split}
\end{equation}
Here $\epsilon_{\mathcal{T}}(h)$, $\epsilon_{\mathcal{S}}(h)$ and $f$ refer to target risk, source risk and labeling function, respectively. 
In a typical \emph{unsupervised DA} setting, it is not possible to measure the third term in Eq.~\ref{Eq: Lemma from Han Zhao}. 
One possible way to measure this term is to query some data labels from target domain so that the learner can learn the conditional relations in the target domain. However, the label annotations usually is expensive. Notice that the convergence rate at the disagreement term would generally be $\mathcal{O}(1/\sqrt{N_t})$ \cite{Mohri-et-al:scheme} with \emph{slow} convergence behaviour if the label is $i.i.d.$ sampled from the target set with size $N_t$, which is far sufficient to minimize the last term. 








To alleviate such difficulties, one can use \emph{Active Learning} (AL) technique for DA so that the learner can reduce the cost of acquiring labels by requesting labeling from the oracle. 
AL only tries to query the labels of the most informative examples, and has been shown, in some optimal cases, to achieve \textbf{exponentially-lower label-complexity} (number of queried labels) than passive learning \cite{cohn1994improving}. 
From this perspective, we tried to break the general $i.i.d.$ sampling with limited information in the target domain (\emph{a.k.a} semi-supervised domain adaptation approach).  
Most previous active learning approaches were rooted in uncertainty-based approaches. \cite{dasgupta2011two} pointed out that only focusing on the uncertainty might lead to \emph{sample bias}. 
To overcome such bias problems, we also need to consider the diversity in the query process. 
Recently, \cite{sinha2019variational,shui2019deep} proposed adversarial training techniques to query the most informative features via a critic function, which could overcome the sample bias problems.


Aiming to address all the aforementioned issues, we proposed a three-stage discriminative active domain adaptation algorithm, which aims to actively query the most informative instances in the target domain to minimize the labeling disagreement term, under the same and small querying label budget. 

In the first stage, we adopted the Wasserstein Distance-based adversarial training technique for unsupervised DA through training a critic function for learning the domain invariant feature. The critic could also be used to discriminate the target domain features for active querying. 
In the second stage, we derived a sample-efficient and straightforward active query strategy based on the network structure, for sampling the most \emph{informative} samples in the target domain by controlling \emph{uncertainty} and \emph{diversity} for selecting the target instances. 
Finally in the third stage, we deployed a re-weighting technique based on the prediction uncertainty for determining the importance of queried samples to retrain the network. 

We then implemented extensive experiments on four benchmark datasets. The empirical results showed that our proposed algorithm could improve the classification accuracy with a small query budget. When the query budget is small, the proposed approach can have better performance than its $i.i.d$ (random) selection counterparts (reported in Table~\ref{tab:Different query budge}), which confirms the effectiveness of our algorithm.  


\section{Related Works}
\paragraph{Domain Adaptation}
A large number of efforts have been addressed toward DA \cite{wang2018deep}. As stated before, many of the previous advancements \cite{ben2010theory,ganin2016domain,tzeng2017adversarial,shen2017wasserstein} were based on the assumption that the conditional relations remain unchanged during the adaptation process. Some recent works proposed to tackle the conditional shifts problem. \cite{long2018conditional} adopted the Conditional Generative Adversarial Nets (CoGANs) to extract the cross-covariance between the source and target feature representations, and also measure the conditional entropy as an uncertainty measure to control the transferability. \cite{wen2019bayesian} proposed the Bayesian Neural Network with entropy and variable uncertainty measures to jointly match the marginal distribution ($\Proba(\x)$) and conditional distribution ($\Proba(y|\mathbf{x})$).

\paragraph{Active Learning}
AL has been widely investigated by academia in the context of theory or applications. 
Recently, \cite{sinha2019variational} proposed a variational autoencoder based adversarial approach to query the informative unlabeled feature from the labeled ones and  \cite{gissin2019discriminative} proposed discriminative active learning. \cite{shui2019deep} extended and adopted a critic network for querying the diverse features. 
Those above usually assumes that labeled and unlabeled data are from same distribution. Few works were proposed to implement active learning for enhancing domain adaptation $i.e.,$ two or more distributions. 

\paragraph{Active Learning for Domain Adaptation}
\cite{persello2012active} proposed a two-direction AL algorthim for DA: query the most informative from the target domain and remove the most strange features out of the source domain. \cite{pmlr-v32-wangi14} proposed the active transfer technique for the model shift problem while assuming the shifts are smooth and implemented conditional distribution matching algorithm and off-set algorithm to modelling the source and target tasks via comparing the Gaussian Distributions. \cite{zhang2013domain} proposed a distribution correction algorithm over kernel embeddings to handle the target shift. The last two methods held on the assumption that there existed an affine transformation of conditional distribution from the source to target. 
\cite{Su_2019_CVPR_Workshops} proposed an active learning method using $\mathcal{H}$ divergence and the importance sampling technique to query the target instances. However, the importance sampling, query strategy they adopted, assumed that $\text{supp}(\mathcal{T})\subseteq \text{supp}(\mathcal{S})$, may not hold in many DA settings.

\section{Problem Setup} \label{sect:problem setup}
\paragraph{Notations and Basic Definitions}
We consider a classification task, denote $\mathcal{X}$ and $\mathcal{Y}$ as the input and output space. A learning algorithm is then provided with a \emph{labeled source dataset} $S=\{(\mathbf{x}_i,y_i)\}^{m_s}_{i=1}$ consisting of $m_s$ examples drawn $i.i.d.$ from $\mathcal{S}_{\x\times y}\sim\mathcal{D}_s$ and an unlabeled target sample $T=\{\mathbf{x}_j\}_{j=1}^{m_t}$ consisting of $m_t$ examples drawn $i.i.d.$ from $\mathcal{T}_\mathbf{x}$, where $\mathcal{S}_{\x\times \y}$ is the joint distribution on $\x\times y$ and $\mathcal{T}_\mathbf{x}$ is the marginal target distribution on $\mathbf{x}$, respectively. 
The expected source and target risk of $h\in\mathcal{H}$ over $\mathcal{S}$ (respectively, $\mathcal{T}$), are the probabilities that $h$ errs on the entire distribution $\mathcal{D}_S$ (respectively, $\mathcal{D}_T$): 
        $\epsilon_{\mathcal{S}}(h)=\mathbb{E}_{(\mathbf{x},y)\sim\mathcal{S}}\mathcal{L}(h(\mathbf{x},y))$ and
$\epsilon_{\mathcal{T}}(h)=\mathbb{E}_{(\mathbf{x},y)\sim\mathcal{T}}\mathcal{L}(h(\mathbf{x},y))$,
where $\mathcal{L}(\cdot)$ is the loss function. The goal of DA is to build a classifier $h\in \mathcal{H}: \mathcal{X}\rightarrow \mathcal{Y}$ training on source domain with a low \emph{target risk} $\epsilon_{\mathcal{T}}(h)$.

\subsection{Optimal Transport and Wasserstein Distance}

Optimal Transport (OT) theory and Wasserstein Distance were recently widely investigated in machine learning \cite{arjovsky2017wasserstein} especially in the domain adaptation area \cite{courty2016optimal}. We follow \cite{redko2017theoretical} and define $c:\calX\times\calX \to \R^{+}$ as the cost function for transporting one unit of mass $\x$ to $\x'$, then Wasserstein Distance could be computed by
\begin{equation*}
W_p^p(\D_i,\D_j) = \inf_{\gamma\in \Pi(\D_i,\D_j)} \int_{\calX\times\calX} c(\x,\x')^p d \gamma(\x,\x')
\end{equation*}
where $\Pi(\D_i,\D_j)$ is joint probability measures on $\calX \times \calX$ with marginals $\D_i$ and $\D_j$ referring to all the possible coupling functions. Throughout this paper, we shall use Wasserstein-1 distance only ($p=1$). 
According to \emph{Kantorovich-Rubinstein} theorem, let $f$ be a Lipschiz-continous function $||f||_L < 1$, we have
\begin{equation}
    W_1(\mathcal{D}_i,\mathcal{D}_j) = \sup_{||f||_L < 1} \mathbb{E}_{x\in\mathcal{D}_i}f(x) - \mathbb{E}_{x^\prime \in \mathcal{D}_j}f(x^\prime)
\end{equation}{}

\subsection{Conditional Shift and Error Bound}
From a probabilistic perspective, the general learning process of most previous DA approaches is to learn the joint distribution of the target domain $\Proba_T(\mathbf{x},y)$ through source domain joint distribution $\Proba_S(\mathbf{x},y)$. Note that  
$\Proba_T(\mathbf{x},y) = \Proba_T(\mathbf{x}|y)\Proba_T(\mathbf{x})$, to guarantee a successful transfer from source domain $S$ to target domain $T$, the underlying assumption is $\Proba_S(y|\mathbf{x})\approx \Proba_T(y|\mathbf{x})$. Recently, \cite{wen2019bayesian} showed that such condition is not sufficiently hold. 


For the conditional shift situation, $\mathbb{P}_S(y|\mathbf{x})\neq \mathbb{P}_T(y|\mathbf{x})$. \cite{zhao2019learning} theoretically showed that such a conditional shift problem exists in many situations and that typically if we only try to minimize the source error together with the domain distances, the target error might increase, which shall hinder the adaptation process. Their analysis was based on $\hat{\mathcal{H}}$ divergence, which is somehow hard to compute in deep learning based methods. 
In order to be coherent with our proposed work, we shall present it using Wasserstein Distance with the following Theorem \ref{Thm:Conditional shift bound}.

\begin{theorem}\label{Thm:Conditional shift bound}
         Let $\big<\mathcal{D}_{\mathcal{S}}, f_{\mathcal{S}}\big>$ and $\big<\mathcal{D}_{\mathcal{T}}, f_{\mathcal{T}}\big>$ be the source and target distributions and corresponding labeling function, if the hypothesis $h$ is 1-Lipschtiz and the loss function is $0-1$ loss, then we have
        \begin{equation}\label{Eq.conditional_shift_bound}
        \epsilon_{\mathcal{T}}(h)\leq\epsilon_\mathcal{S}(h)+2W_1(\D_s,\D_t) +\mathbb{E}_{\mathcal{D}_\mathcal{S}}\big[\big|f_\mathcal{S}-f_\mathcal{T}\big|\big]
        \end{equation}{}
\end{theorem}
The proof is based on Lemma 1 of \cite{shen2017wasserstein} and is symetric to the proof of Theorem 3 of \cite{zhao2019learning}. Due to space limit, we show the sketch idea of proof,
\begin{proof}
Based on Lemma 1 of \cite{shen2017wasserstein}, let $h^{\prime} = f_T$, we have 
\begin{equation*}
    \epsilon_t(h,f_T)\leq\epsilon_s(h,f_T)+2W_1(\D_s,\D_t)
\end{equation*}
We noticed that 
\begin{equation*}
    \begin{split}
        \epsilon_s(h,f_T) & = \E_{x\sim\D_s}|h(x)-f_T(x)|\\
        & \leq \E_{x\sim\D_s}|h(x)-h_S(x)| + \E_{x\sim\D_s}|h_S(x)-f_T(x)| \\
        & = \epsilon_s(h) + \E_{x\sim\D_s}|h_S(x)-f_T(x)|\\
    \end{split}
\end{equation*}
Plugging in we have the result.
\end{proof}{}
This theorem showed that error on the target domain is decided by source domain error, Wasserstein Distance between source and target, and the conditional distribution on both source and target domains. Here the third term is not measurable in the unsupervised domain adaptation setting. If the conditional distribution changes during the adaptation process, then the target error may diverge~\cite{zhao2019learning}. One direct approach to reduce the disagreement between $f_\mathcal{S}$ and $f_\mathcal{T}$ is to partially acquire the labeling function $f_\mathcal{T}$, i.e., the labels in the target domain. 

Besides, the Wasserstein distance between the source and target distribution (second term in Eq.~\ref{Eq.conditional_shift_bound}), is measured by total transportation cost between the source domain. Denote $\mathcal{D}_U$ and $\mathcal{D}_L$ by the corresponding distributions of unlabeled and labeled datasets, then the Wasserstein distance is denoted by:
\begin{equation*}
W_1(\mathcal{D}_U,\mathcal{D}_L) = \inf_{\gamma\in \Pi(\mathcal{U},\mathcal{L})} \int_{\calX\times\calX} c(\x_l,\x_u) d \gamma(\x_l,\x_u)
\end{equation*}

Intuitively, if we can query some instances in the target domain $\mathcal{T}$ ($\mathcal{D}_U$) and move them from target into the source domain $\mathcal{S}$ $(\mathcal{D}_L)$, we can reduce the total transportation cost between the two domains, $i.e., $ the Wasserstein distance between the two domains. 

Based on this, to minimize the RHS of Eq.~\ref{Eq.conditional_shift_bound} is equivalent to train a learner $h\in\mathcal{H}$ that: $1)$ minimize the source error; $2)$ train a critic to estimate the empirical Wasserstein Distaince between the source and target domain and approximately find a feature extractor that can minimize the total transportation cost between the source and target domain in an adversarial way with the critic; $3)$ can query the labeling information in the target domain so that to minimize the disagreement of labeling function between the source and target domain $i.e.,$the third term of Eq.~\ref{Eq.conditional_shift_bound}. 

To this end, we argue that if the learner can actively query labeling information in the target domain, then, it can partially get the conditional information in the target domain. 
With the minority of labeled target instances in hand, it can learn to jointly minimize the error both on the source and target domain. 
Furthermore, to $i.i.d.$ query the label is somehow slow. In order to reduce the annotation expense, we may expect the learner to query some informative instances using an active learning strategy. 
Also, if the queried instances in the target domain are informative enough, they will have a better representative property on the target domain. 
Then, the learner can have better generalization performance on the target domain. Take those above into consideration, we can formally propose the discriminative active domain adaptation method.

\section{Active Discriminative Domain Adaptation}

Our learning process mainly consists of three main stages. We will introduce them in details. 

\subsection{Stage 1: Domain Adversarial Training via Optimal Transport}
 
For the first stage, we adopt \emph{Wasserstein Distance Guided Representation Learning} \cite{shen2017wasserstein} method for adversarial training. The network receives a pair of instances from the source and target domain. Denoted by $F$ and $C$ the feature extractor and classifier, parameterized by $\theta_f$ and by $\theta_c$, respectively. The feature extractor is trained to learn invariant features, and the classifier is expected to learn the conditional prediction relations $\mathbb{P}(Y|\mathbf{X})$ for predicting the instances from both source and target domain correctly. For the classification loss, we employ the traditional cross-entropy loss:
 $\mathcal{L}_{cls} = -\sum_{i=1}^m y_i\log(\mathbb{P}(C(F(\textbf{x}_i))))$.

Then, there follows the domain critic network $D$, parameterized by $\theta_d$. 
It estimates the empirical Wasserstein Distance between the source and target domain through a pair of batched instances $\mathcal{X}_\mathcal{S}$ and $\mathcal{X}_\mathcal{T}$,
\begin{equation}
    \begin{split}
            W_1(\mathcal{X}_\mathcal{S},\mathcal{X}_\mathcal{T})=\frac{1}{n_s}&\sum_{\x_s\in\mathcal{X}_\mathcal{S}}D(F(\x_s))-\frac{1}{n_t}\sum_{\x_t\in\mathcal{X}_\mathcal{T}}D(F(\x_t))
            \label{Eq.Wasserstein}
    \end{split}{}
\end{equation}


The feature extractor $F$ is then trained to minimize the estimated Wasserstein Distance in an adversarial manner with the critic $D$. Then, goal of first stage training is described by
\begin{equation}
    \min_{\theta_f,\theta_c}\max_{\theta_d} \mathcal{L}_{cls} + \lambda_w(W_1(\mathcal{X}_\mathcal{S},\mathcal{X}_\mathcal{T}) - \mathcal{L}_{grad})
    \label{Eq.objective1}
\end{equation}{}
where $\lambda_w$ is a trade-off coefficient and $\mathcal{L}_{grad}$ is the gradient penalty term suggested by \cite{gulrajani2017improved}.
The source and target features (marginal distributions) could be aligned via such an adversarial training process. Then, based on this aligned marginal distribution, we can implement the active strategy to query the most informative target instances

\subsection{Stage 2: Active Query with Wasserstein Critic}

\begin{figure}[tb]
		\centering 
		\includegraphics[width=0.45\textwidth]{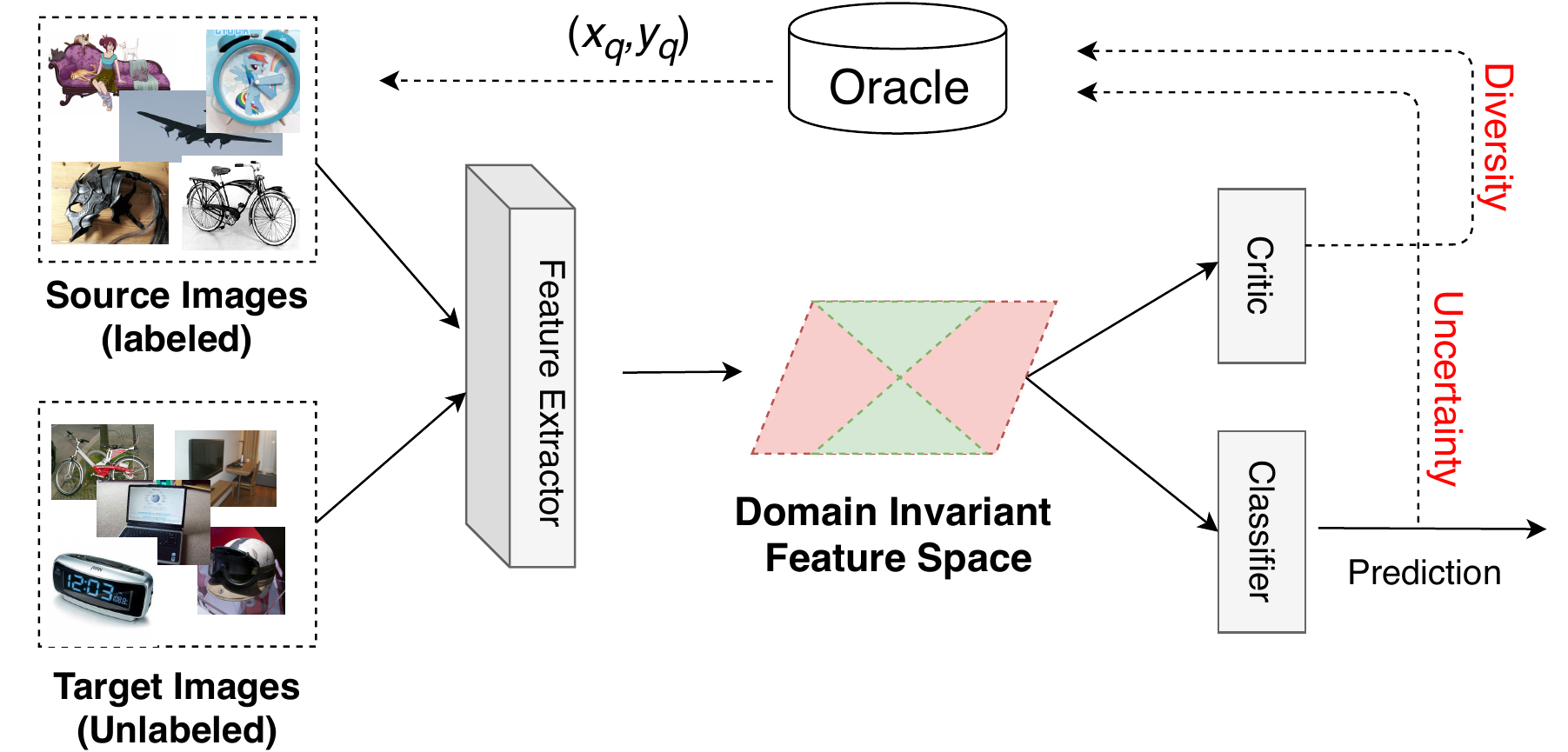}
\caption{Ac-DA workflow: feature extractor are trained to learn a domain invariant feature space together with the critic. The learner selects the informative instances by measuring \emph{uncertainty} and \emph{diversity} based on critic and classifier outputs. } 
\label{fig:stage2}
\end{figure}

For the second stage, we hope the active leaner can find out the most informative features among the unlabeled target so that it could leverage from the labeling information of the target domain. 
The informative features, intuitively, are the \emph{ones most different from what the learner has already known}. 
Intuitively, the hardest instances to adapt are those with least confidence, $i.e.$ the most uncertain ones, to predict based on current classifier. 
As pointed out as previous work~\cite{dasgupta2011two}, only focus on the uncertainty shall lead to the \emph{sampling bias}. In order to reduce the sampling bias, the active learner shall also search diversity some target samples. We therefore find the most informative target samples holding both uncertainty and diversity properties. 


\paragraph{Prediction Uncertainty}
The conditional prediction $\Proba_T(Y|\mathbf{X})$ is learned by the classification network. To measure the uncertainty, we can borrow the idea from \cite{long2018conditional} to adopt entropy measure to quantify the uncertain of the classifier. The uncertainty entropy measure over an instance $\mathbf{x}_t$ is denoted by
\begin{equation}
    \mathcal{U}(y_t|{\mathbf{x}}_t) = \mathcal{H}(\hat{\mathbb{P}}(y_t|{\mathbf{x}}_t))
    \label{Eq:Uncertainty}
\end{equation}

where $\mathcal{H}(\cdot)$ is the information entropy measure, $\hat{\mathbb{P}}(y_t|\mathbf{x}_t)$ is the output of classification network $\hat{\mathbb{P}}(y_t|\mathbf{x}_t)=C(F(\mathbf{x}_t))$.
\paragraph{Diversity by Critic Function}
If some instances, in terms of distribution distance measures, are very far from the unknown labeled ones, then they should contain most informative and diverse features from the known labeled ones. 
Recall that in the first stage, we match the marginal distribution between the source and target domain to achieve a domain invariant feature space with Wasserstein Distance. 
Then, for the target domain instances, the one with highest critic score is the one that have the highest transportation cost.

\cite{sinha2019variational,shui2019deep} showed that such critic term $D(F(\cdot)):\calX\to[0,1]$ indicates the diversity in the query process. Then, we can leverage from the trained Wasserstein Critic network to evaluate and find out the most informative (diverse) target features on the invariant feature space. That is, measuring the diversity of target instances via critic score. Consider the critic output of a target instance $\x_t$, 
if $D(F(\x_t))\to 1$, then $\x_t$ is far, $w.r.t.$ Wasserstein Distance, from the source domain images and if $D(F(\x_t))\to 0$, then $\x_t$ is near to the source images.

Based on those above, if we hope to find out the most informative (uncertain and diverse) instances in the target domain, then we should query by controlling two terms:  
\begin{itemize}
    \item uncertainty score $\mathcal{U} =\mathcal{H}(\hat{\mathbb{P}}(y_t|{\mathbf{x}}_t)) $ defined by Eq.~\ref{Eq:Uncertainty}, which is indicates the uncertainty of the classifier to predict a label $y^?_t$ given the instance $\mathbf{x}_t$
in the target domain
   \item critic score $D(F(\x_t))$ by the the Wasserstein critic function, which indicates the diversity of the unlabeled target instance compared with the source labeled ones.
\end{itemize}
Then, we shall have the following objective
\begin{equation}
     \mathrm{argmax}_{x_t\in\mathcal{X}_t} \mathcal{U}(y^?_t|\mathbf{x}_t)-\lambda_{div}D(F(\x_t))
  \label{Eq.query_eq}
\end{equation}
where $\lambda_{div}$ is a coefficient to regularize the Wasserstein critic term. So, for a query budget $\beta$ and $m_t$ of target set instances, the query process could be described as: \emph{looking for $m_q = \beta m_t$ instances by solving Eq.~\ref{Eq.query_eq} and query the labels of those $m_q$ instance from the oracle}. Denote the queried set by $Q=\{(\mathbf{x}_1^q,y^q_1), \dots,(\mathbf{x}_{m_q}^q,y^q_{m_q})\}$. Then, uniting such small batch instances with the source domain and removing them from the target domain. The source and target datasets shall be updated as: $S^\prime=S\cup Q$, $T^\prime =T/Q$.
We illustrate a general query workflow in Fig.~\ref{fig:stage2}. 

\begin{algorithm}[tb]
\caption{The Active Discriminative Domain Adaptation}
\label{alg:algorithm}
\textbf{Input}: Source and target domain input $S$, $T$; Query budget $\beta$\\
\textbf{Parameter}: Feature extractor $\theta_f$; Classifier $\theta_c$; Critic $\theta_d$ \\
\textbf{Output}: Optimized $\theta_f^\star$, $\theta_c^\star$, $\theta_d^\star$
\begin{algorithmic}[1] 
\WHILE{Domain level adaptation not finish}
\STATE Sample batches $(\x_s,y_s)\sim S$, $\x_t\sim T$
\STATE Train the network based on Eq.~\ref{Eq.objective1} until converge
\ENDWHILE
\IF {Query budget is not empty}
\STATE Select the target instances $ \{\mathbf{x}_1^q, \dots,\mathbf{x}_{m_q}^q\}$ according to Eq.~\ref{Eq.query_eq} and query the label $\{y_1^q,\dots,y_{m_q}^q\}$ from oracle. 
\ELSE 
\STATE Update the dataset $Q=\{(\mathbf{x}_1^q,y^q_1),\dots,(\mathbf{x}_{m_q}^q,y^q_{m_q})\}$, $S^\prime=S\cup Q$, $T^\prime =T/Q$.
\ENDIF
\STATE Compute the uncertainty vector $\balpha = [\alpha_1,\dots,\alpha_C]_{j=1}^C$ with Eq.~\ref{uncertainty_vector}
\STATE Train the network on new labeled and unlabeled dataset via domain adaptation techniques with Eq.~\ref{Eq.stage3}.
\STATE \textbf{return} solution
\end{algorithmic}
\end{algorithm}

\subsection{Stage 3: DA training with new dataset}
The goal of our proposed method is to leverage the most informative instances in the target domain to reinforce the adaptation process. 
General adversarial training methods for domain adaptation usually assign each instance with the same importance weight. In order to enforce the uncertainty information to the classifier, we hope to give higher weights to the instances with higher uncertainty scores during the supervised classification process. 

Denote by a set of $m_q$ queried instances $\{\mathbf{x}^{(i)}, y^{(i)}\}_{i=1}^{m_q}$, we shall re-weight the importance of each instance classes based on their uncertainty score. Denote by uncertainty vector $\balpha = [\alpha_1\dots\alpha_j\dots\alpha_C]_{j=1}^C$ over all $C$ classes. For each class $j$, the weight is computed by,
\begin{equation}
\alpha_j = \frac{N_j\cdot \mathcal{U}(y_j|\mathbf{x})}{\sum_{i=1}^{m_q}\mathcal{U}(y^{(i)}|\mathbf{x})}
\label{uncertainty_vector}
\end{equation}
where $N_j$ is the number of instances with label $y_j$, $\mathcal{U}(\cdot)$ is the uncertainty score defined in Eq.\ref{Eq:Uncertainty}.

For a batch of queried instances, the weighted crossentropy loss could be computed by
$$\mathcal{L}^q_w = \alpha_j(-y_j\log(\sum_{j=1}^C\exp(\Proba(y_j|\mathbf{x}))))$$

Then, objective function for the third stage is,
\begin{equation}
        \min_{\theta_f,\theta_c}\max_{\theta_d} \mathcal{L}^q_w +\mathcal{L}_{cls}+ \lambda_w(W_1(\mathcal{X}_\mathcal{S}^\prime,\mathcal{X}_\mathcal{T}^\prime) - \mathcal{L}_{grad})\label{Eq.stage3}
\end{equation}{}
where $\mathcal{X}_\mathcal{S}^\prime$ and $\mathcal{X}_\mathcal{T}^\prime$ are sampled from the updated source and target datasets, $\mathcal{L}_{cls}$ is the classification loss on the original source set and $\mathcal{L}^q_w$ is the weighted loss for the query set.
Finally, we illustrate our Active Discriminative Domain Adaptation (Ac-DA) algorithm in Algorithm~\ref{alg:algorithm}

\section{Experiments and Results}
 We evaluate the performance of the proposed algorithm on four benchmark datasets and compared with some other approaches: Wasserstein Guided Domain Adaptation (WDGRL~\cite{shen2017wasserstein}), Domain Adversarial Neural Networks (DANN~\cite{ganin2016domain}), Adversarial Discriminative Domain Adaptation (ADDA~\cite{tzeng2017adversarial}) and Conditional Adversarial Domain Adaptation (CDAN~\cite{long2018conditional}). In order to show the benefits of active query method, we also compare the results with random selection process when the query budget is the same. All experiments are programmed by \emph{Pytorch}. 

\subsection{Datasets and Implementations}
We test our proposed algorithm on four benchmark datasets.
\paragraph{Digits Datasets}
We test our algorithm on digits datasets with the experiments setting : USPS (U)$\leftrightarrow$ MNIST (M) and MNIST $\to$ MNIST-M (MM). 
For USPS we resize the images to size $28\times 28$. We train the network using training sets with size: MNIST/MNIST-M($60k$), USPS($7,291$) and testing sets with size: MNIST/MNIST-M ($10k$), USPS($2,007$). 



\begin{table}
\centering

\resizebox{0.40\textwidth}{!}{\begin{tabular}{l|ccccc}  
\toprule
Method  & M $\to$ MM & M $\to$ U & U $\to$ M    & avg. \\
\midrule
LeNet5 & $56.1$ & $67.4$ & $65.3$ &  $60.3$\\
DANN       & $74.2$  & $77.1$   &  $73.2$   & $74.6$\\
WDGRL   & $80.3$  & $81.1$ & $74.2$ &   $78.5$    \\
ADDA & $78.9$& $83.5$& $82.3$& $81.5$\\
\midrule
Rand.    & $92.4$  & $\mathbf{95.7}$ &  $95.8$& $94.7$\\
Ac-DA   &  $\mathbf{95.4}$ & $95.5$  &  $\mathbf{96.5}$ &  $\mathbf{95.6}$ \\
\bottomrule
\end{tabular}}
\caption{Classification accuracy ($\%$) on \textbf{digits datasets} with different adaptation tasks. The last two line are our method, Random refers to randomly query some instance while Ac-DA is the proposed approach. Both two methods are restrict to $10\%$ query budget. }
\label{tab:digits}
\end{table}

\begin{table}[tb]
\centering
\resizebox{0.45\textwidth}{!}{
\begin{tabular}{l|cccccc}  
\toprule
Method  & A $\to$ W  & A $\to$ D & D $\to$ A  & W $\to$ A & avg. \\
\midrule
{ResNet50}& $68.6  $   & $69.3$ & $61.1$ & $60.7$ & $64.9$ \\
DAN  & $80.5  $   & $78.6$ & $63.6$ & $60.7$ & $62.7$ \\
DANN       & $81.3$  &   $79.2$ & $68.2$ & $67.4$ & $74.0$\\
WGDRL            & $79.2$  & $80.2$ & $69.3$& $69.1$ & $74.5$    \\

\hline
Rand.  & $86.1$  &  $85.6$ & $ 76.3$ & $ 78.1 $ & $81.6$\\
Ac-DA  &  $\mathbf{86.6}$   & $\mathbf{87.7}$ & $\mathbf{78.5}$ & $\mathbf{80.2}$ &  $\mathbf{83.3}$\\
\bottomrule
\end{tabular}}
\caption{Classification accuracy ($\%$) on \textbf{Office-31} dataset with different adaptation settings with $10\%$ query budget. }
\label{tab:office_results}
\end{table}

\begin{table*}[tb]
\centering
\resizebox{1\textwidth}{!}{
\begin{tabular}{l|ccccccccccccc}  
\toprule
Method  & Ar $\to$ Cl  & Ar $\to$ Pr & Ar $\to$ Rw  & Cl $\to$ Ar & Cl $\to$ Pr  & Cl $\to$ Rw & Pr $\to$ Ar & Pr $\to$ Cl & Pr $\to$ Rw & Rw $\to$ Ar& Rw $\to$ Cl & Rw $\to$ Pr& avg. \\
\midrule
{ResNet50}& $ 34.9$   & $50.0$ & $58.0$ & $37.4$ & $41.9$ &$46.2$&$38.5$&$31.2$&$60.4$& $53.9$ & $41.2$& $59.9$&$46.1$\\
DANN       & $ 45.6$   & $59.3$ & $70.1$ & $47.0$ & $58.5$ &$60.9$&$46.1$&$43.7$&$68.5$& $63.2$ & $51.8$& $76.8$&$57.6$\\
WGDRL            & $42.6$  & $57.9$ & $69.3$& $47.3$ & $59.5$ &  $63.4$ &$46.2$ & $41.3$ &$67.4$ &$62.4$ & $52.8$ & $74.9$ & $57.1$\\
CDAN &$47.2$ & $62.5$& $72.6$ & $51.8$ & $62.2$ & $66.1$ & $51.4$ & $46.3$ & $70.1$ & $66.3$ & $53.1$ & $78.7$ & $60.7$\\

\hline
Rand.   & $56.9$  &  $76.4$ & $ 76.3$ & $ \mathbf{61.7} $ & $78.1$ & $73.3$ & $57.8$& $\mathbf{56.9}$& $74.2$ &$68.5$& $60.3$ & $83.2$ & $68.6$\\
Ac-DA  &  $\mathbf{57.5}$   & $\mathbf{76.9}$ & $\mathbf{80.2}$ & $61.1$ &  $\mathbf{78.4}$ & $\mathbf{76.7}$ & $\mathbf{59.2}$ & $56.2$ & $\mathbf{79.6}$ & $\mathbf{75.4}$ & $\mathbf{62.5}$ & $\mathbf{85.3}$  &$\mathbf{70.7}$\\
\bottomrule
\end{tabular}}
\caption{Classification accuracy ($\%$) on \textbf{Office Home} dataset with different adaptation settings with query budget $10\%$.}
\label{tab:office_home_results}
\end{table*}

\begin{table}
\centering

\resizebox{0.48\textwidth}{!}{\begin{tabular}{l|ccccccc}  
\toprule
Method       & C $\to$ I & C$\to$P & I $\to$ P & I $\to$ C & P $\to$ C & P $\to$ I & avg. \\
\midrule
ResNet50     & $76.4$   & $62.5$   & $73.2$   & $89.3$  &$90.3$ & $79.8$  & $78.5$\\
DANN         & $84.8$   & $72.6$   & $73.8$   & $92.8$  &$91.5$ & $81.9$  & $82.9$\\
WDGRL        & $82.3$   & $70.8$   & $73.9$   & $90.7$  &$91.3$ & $85.4$  & $82.4$ \\
CDAN         & $87.5$   & $73.4$   & $75.3$   & $93.1$  &$92.8$ & $87.2$  & $84.8$\\
\midrule
Rand.        &  $89.8$  & $75.0$   & $78.2$   & $94.4$  &$\mathbf{94.9}$ & $89.9$  &$87.1$\\
Ac-DA    &  $\mathbf{91.1}$ & $\mathbf{76.3}$&  $\mathbf{80.8}$ & $\mathbf{96.7}$  &$94.7$& $\mathbf{94.2}$ & $\mathbf{88.9}$ \\
\bottomrule
\end{tabular}}
\caption{Classification accuracy ($\%$) on \textbf{Image-CLEF dataset} with different adaptation tasks under $10\%$ query budget. }
\label{tab:image_clef}
\end{table}



\paragraph{Office-31 dataset} is a standard benchmark for domain adaptation evaluations. It contains three different domains: Amazon (A), Dslr (D) and WebCam (W), with $31$ categories in each domain. We report the average results in Table~\ref{tab:office_results}. 
\paragraph{Office Home dataset} is more challenging than Office-31, contains four different domains: \emph{Art} (Ar), \emph{Clipart} (Cl), \emph{Prodcut} (Pr) and \emph{Real World} (Rw), with $65$ categories in each domain. We report the average results in Table~\ref{tab:office_home_results}.
\paragraph{Image-CLEF 2014 dataset} contains three domains, which are \emph{Caltech-256}(C), \emph{ILSVRC-2012}(I), and PascalVOC-2012(P), with 12 common shared catagories. We report the average results in Table.~\ref{tab:image_clef}
 
For digits datasets, we do not apply any data-augmentation. For Office-31, Office-Home and Image-CLEF datasets, we apply the following pre-processing pipline: $1)$ for training set, firstly resize the image to $256\times 256$ then, apply $RandomCrop$ downgrade the size to $224 \times 224$, after that, apply the same random flipping strategy of \cite{you2019universal}; $2)$ for testing set, resize the images to $256 \times 256$ then use \textit{CenterCrop} to size $224 \times 224$. 

\paragraph{CNN Archiecture and Implementations}
\begin{figure}[tb]
		\centering 
		\includegraphics[width=0.23\textwidth]{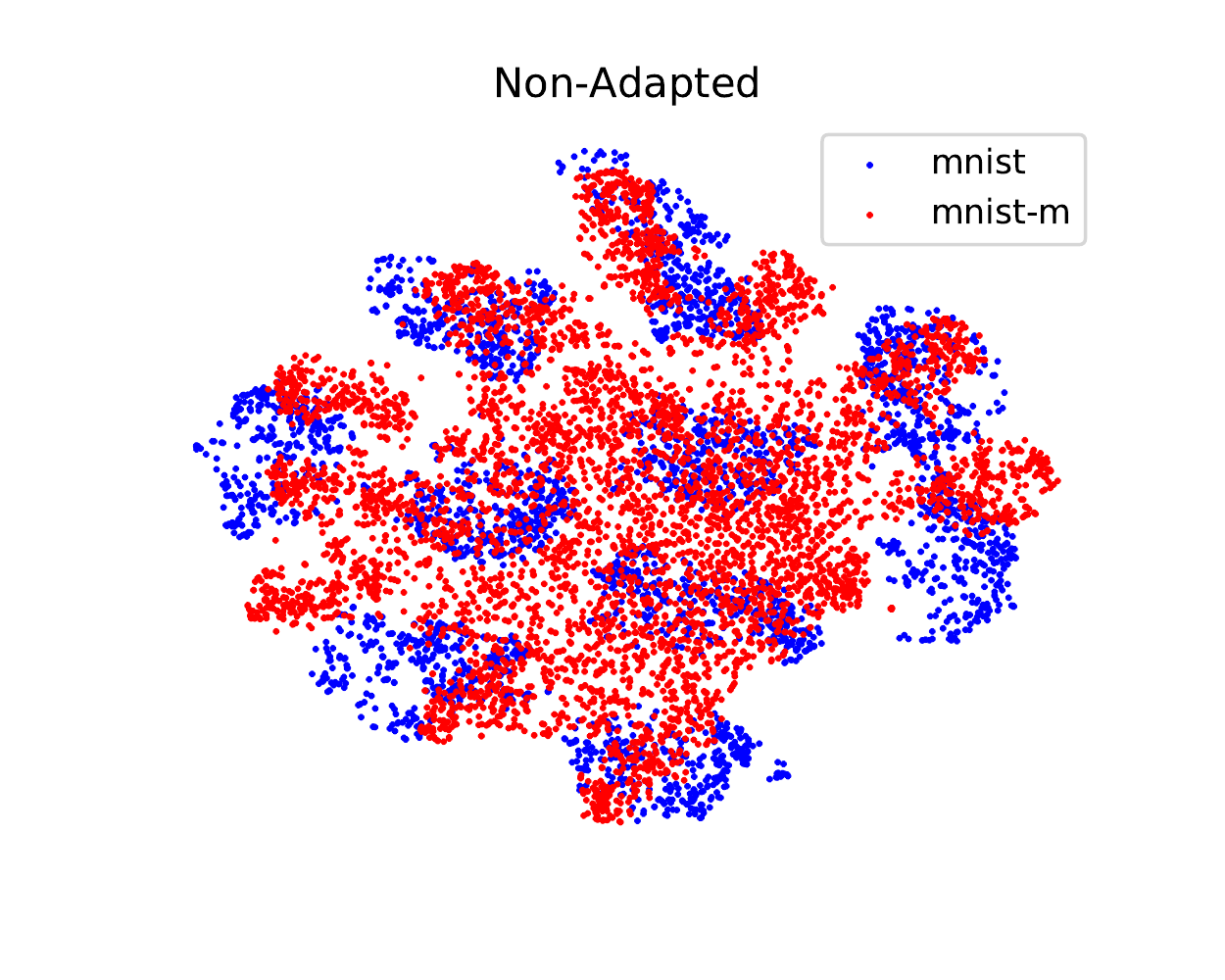} 
		\includegraphics[width=0.23\textwidth]{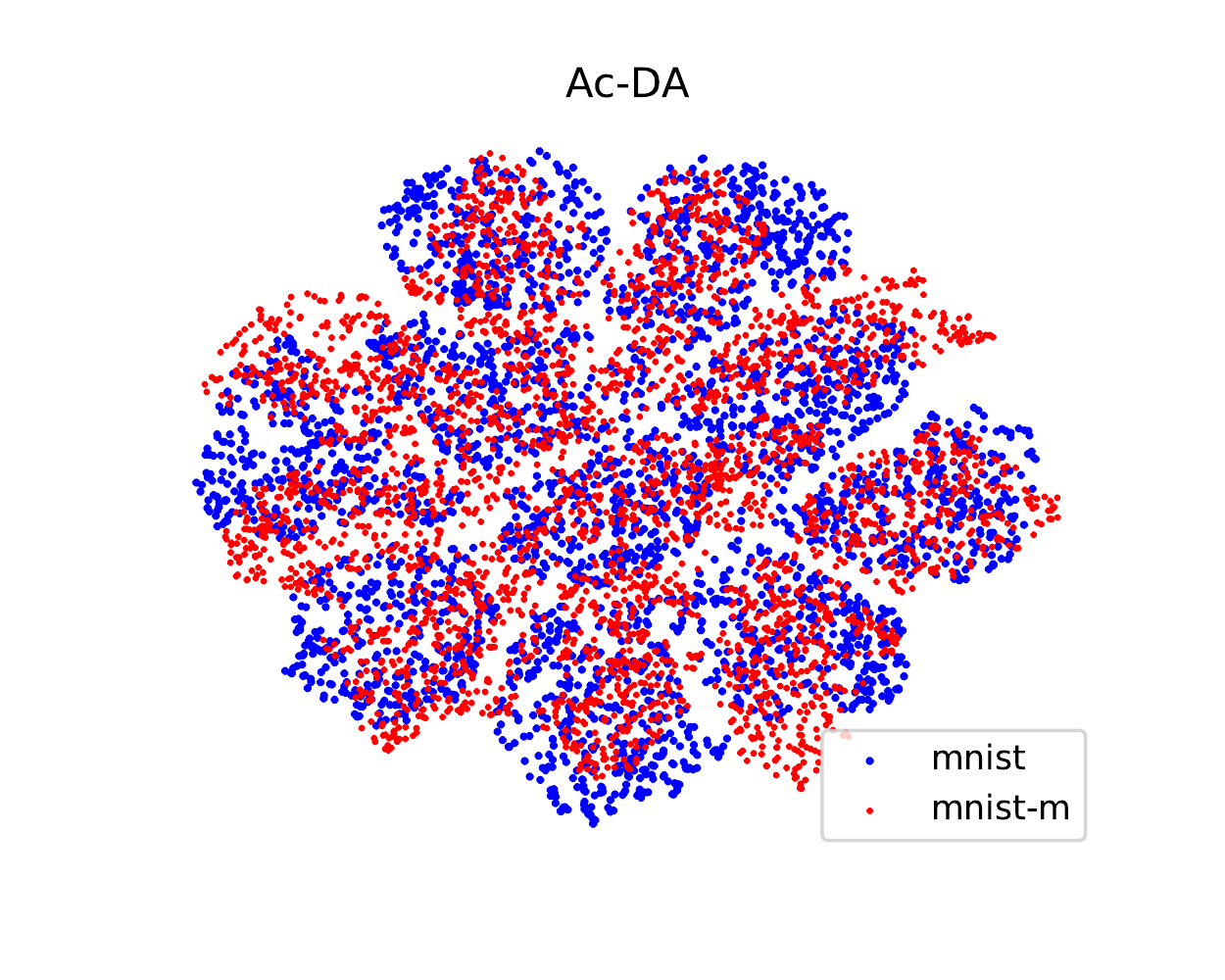}
\caption{T-SNE visualization between our proposed Active Discriminative Domian Adaptation (right, with $5\%$ query budget) and non-adapted setting (left) for MNIST $\to$ MNIST-M adaptation task.} 
\label{fig:tsne}
\end{figure}
For digits experiments, we adopt \emph{LeNet-5} as feature extractor and trained from scratch.
For the rest three real-world datasets, we implement ImageNet pretrained \emph{ResNet-50} as feature extractor. For the digits experiments, we train the network with mini-batch size $64$ and for the rest three datasets with mini-batch size $16$. 
We adopt Adam optimizer for training the network. For stable training, we set $\lambda_{w}= \frac{2}{1+\exp(-\delta p)}-1$, where $\delta=10$ and $p$ is the training progress. Also, we empirically set $\lambda_{div}=10$. To avoid over-training, we also adopt early-stopping technique.

\subsection{Results and Analysis}
We illustrate the T-SNE visualization comparison of non-adaptation setting and our proposed approach Ac-DA. We can observe that our proposed method has a good alignment performance.
We report the average results of our proposed algorithm and baselines using our data pre-processing pipeline on Digits, Office-31, Office-Home and Image-CLEF datasets in Table ~\ref{tab:digits},~\ref{tab:office_results},~\ref{tab:office_home_results} and \ref{tab:image_clef}, respectively.  In order to show the effectiveness of active query strategy, for a given budget, we also implemented \emph{random ($i.i.d.$) selection}  method to query the labels for comparison. The name of such implementations are denoted by \emph{rand.} and \emph{Ac-DA} in each table. In Table~\ref{tab:Different query budge}, we also compared the performances under different budget.



\paragraph{Value of Target Labels}

From the tests results on the four benchmark datasets, we could observe that the to randomly select some instances in the target domain could benefit the classification performance on the target domain. Our method is rooted in WDGRL, comparing accuracy performance between the \emph{random selection} with WDGRL we could observe improvements with $+8.5\%$ on Digits, $+7.6\%$ on Office-31, $+11.5\%$ on Office-Home and $+4.7\%$ on Image-CLEF dataset which confirm the usefulness of label information for adaptation. Also, for each adaptation task on every dataset, we can observe that the proposed Ac-DA algorithm outperforms the random selection method in almost all the tasks. This also confirms that active query can outperform $i.i.d.$ selection.

\paragraph{Effectiveness of Active Query }
We compared the performance with active query and random random selection. We also implement the experiments with different query budgets (with $5\%$, $10\%$ and $15\%$), the average on different dataset is reported in Table~\ref{tab:Different query budge}. we can observe that the accuracy will increase as the query budget increases. Also, for same query budget, we compare the accuracy of active query and random selection. We can observe that active query method can outperform the random query method with query budget $5\%$ and $10\%$. That is, \textbf{with smaller query budget, the active query strategy can have better performance than random selection}. This confirms the effectiveness of active query strategy. 
When the query budget goes to $15\%$, we don't observe distinguishable differences. One interpolation is that as the query budget increase, the more instances in the target domain will be labeled and those most informative ones will be covered with high probability. When the query budget is relatively small, the active strategy can exactly look for the most informative instances rather than uniformly (random) selecting some instances. 



\begin{table}[t]
    \centering
    \resizebox{0.49\textwidth}{!}{
        \begin{tabular}{l|cc|cc|cc}
        \toprule
        & \multicolumn{2}{|c|}{Digits}  & \multicolumn{2}{c|}{Office-Home} & \multicolumn{2}{c}{Image-CLEF}\\ 
         budget &  Rand. & Ac-DA     & Rand. & Ac-DA& Rand. & Ac-DA \\ 
        \midrule
        
        \multicolumn{1}{c|}{$5\%$} & $91.6$   & $92.9(+1.3)$    & $62.4$ & $65.6(+3.2)$ &$82.2$ &$84.9(+2.7)$    \\

        \multicolumn{1}{c|}{$10\%$} & $94.7$  &$95.6(+0.9)$  & $68.6$ & $70.7 (+2.1)$&$87.1$ & $88.9(+1.8)$\\
        
        \multicolumn{1}{c|}{$15\%$} & $96.2$  &  $96.9(+0.7)$    &  $73.9$ & $74.0 (+0.1)$ & $89.8$ & $90.4(+0.6)$  \\
        \bottomrule
\end{tabular}
}
    \caption{Comparison of different query budgets ($5\%$, $10\%$, $15\%$) on three datasets. For each query budget, we report the improvements by applying the active query strategy comparing with the random query strategy in the parentheses.}
    \label{tab:Different query budge}
\end{table}


\section{Conclusion}
We proposed a three-stage discrimative active algorithm to improve the domain adaptation performance. The first stage adopted general domain adversarial training. In the second stage, we proposed an end-to-end query strategy combining \emph{uncertainty} and \emph{diversity} criteria to find out the most informative features in the target domain. Finally, in the third stage, we deployed a re-weighting technique based on the prediction uncertainty for determining the importance of the queried samples to retrain the network. The empirical results confirmed the effectiveness of our active domain adaptation algorithm especially when the query budget is small. 


\bibliographystyle{named}
\bibliography{ijcai20}

\end{document}